\documentclass[10pt,twocolumn,letterpaper]{article}

\usepackage{iccv}              

\newcommand{\norm}[1]{\left\lVert#1\right\rVert}
\DeclareMathOperator*{\argmin}{arg\,min}

\usepackage[accsupp]{axessibility}  
\usepackage[ruled,vlined]{algorithm2e}
\usepackage{algorithmic}
\usepackage{mathtools}
\usepackage[T1]{fontenc}
\usepackage{placeins}
\usepackage{float}
\usepackage{stfloats}

\usepackage{xcolor}

\definecolor{mydarkgreen}{RGB}{0,100,0}

\definecolor{iccvblue}{rgb}{0.21,0.49,0.74}
\usepackage[pagebackref,breaklinks,colorlinks,allcolors=iccvblue]{hyperref}

\def\paperID{14615} 
\def\confName{ICCV}
\def\confYear{2025}

\title{DCT-Shield: A Robust Frequency Domain Defense against Malicious Image Editing}

\author{
    Aniruddha Bala\textsuperscript{1}\thanks{Equal contribution} \quad
    Rohit Chowdhury\textsuperscript{1}\footnotemark[1] \quad
    Rohan Jaiswal\textsuperscript{1} \quad
    Siddharth Roheda\textsuperscript{1} \\
    \textsuperscript{1}Samsung R\&D Institute, Bangalore \\
    {\tt\small \{aniruddha.b, rohit.c, r.jaiswal, sid.roheda\}@samsung.com}
}

\begin{document}
\maketitle



\begin{abstract}
Advancements in diffusion models have enabled effortless image editing via text prompts, raising concerns about image security. Attackers with access to user images can exploit these tools for malicious edits. Recent defenses attempt to protect images by adding a limited noise in the pixel space to disrupt the functioning of diffusion-based editing models. However, the adversarial noise added by previous methods is easily noticeable to the human eye. Moreover, most of these methods are not robust to purification techniques like JPEG compression under a feasible pixel budget. We propose a novel optimization approach that introduces adversarial perturbations directly in the frequency domain by modifying the Discrete Cosine Transform (DCT) coefficients of the input image. By leveraging the JPEG pipeline, our method generates adversarial images that effectively prevent malicious image editing. Extensive experiments across a variety of tasks and datasets demonstrate that our approach introduces fewer visual artifacts while maintaining similar levels of edit protection and robustness to noise purification techniques.
\end{abstract}

\section{Introduction}
\label{sec:intro}

Diffusion models have gained popularity in AI-driven image generation and editing, excelling in tasks such as inpainting, style transfer, and adding or removing objects. In particular, instruction-based image editing has gained momentum, enabling users to modify images using simple text prompts. Models like Stable Diffusion \cite{sd} are now widely available, allowing off-the-shelf use and fine-tuning for specific datasets. However, this accessibility also raises concerns about misinformation and unauthorized content manipulation. Malicious users can exploit these models to alter publicly available images or modify artwork without consent. Thus, developing effective techniques to protect images from unauthorized edits is crucial.

    \begin{figure}[t]
        \centering
        \includegraphics[width=0.48\textwidth]{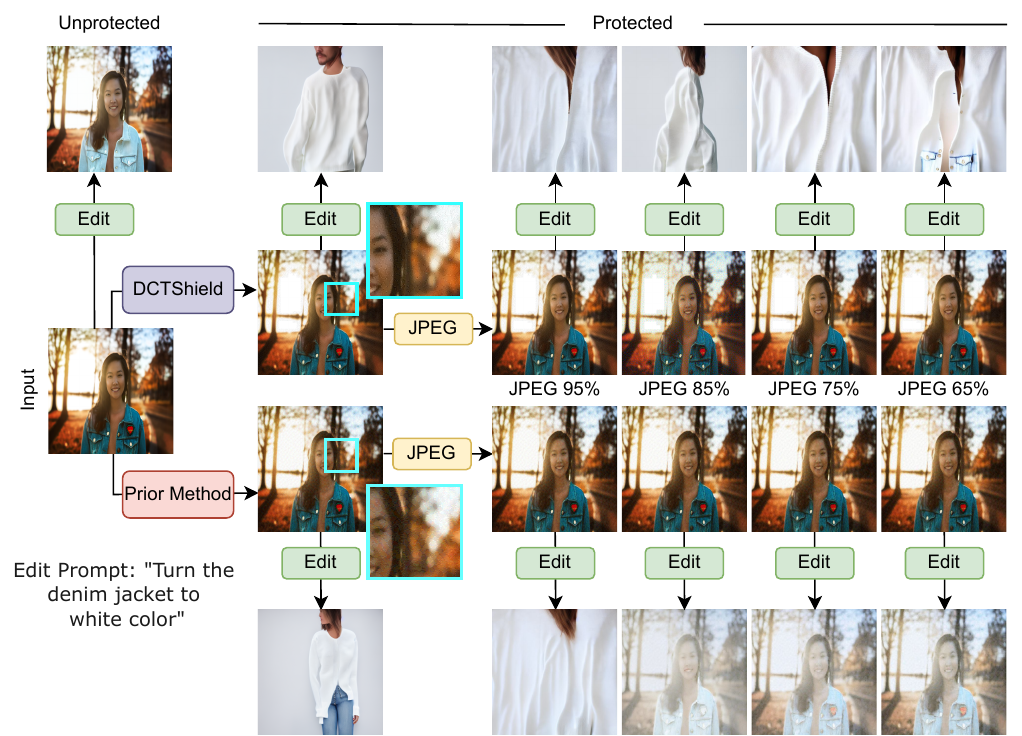}
        \caption{A comparison of our proposed method, DCT-Shield's, performance compared to SDS(-) \cite{xue2024_diffprotect_sds}. Our method produces highly imperceptible adversarial noise while offering strong protection against edits. Immunized images from DCT-Shield are significantly more robust to purification techniques like JPEG-conversion across a wide range of compression qualities.}
        \label{fig:teaser}
    \end{figure}
    
Current approaches protect images by adding small adversarial noise that disrupts diffusion models, preventing meaningful edits. However, existing methods have key limitations that make them impractical. The added noise is often noticeable when zoomed in, and protection fails when images are converted to low-quality JPEG format. As shown in Fig.~\ref{fig:teaser}, prior methods like SDS(-) \cite{xue2024_diffprotect_sds} leave visible artifacts and lack robustness against JPEG compression.

In this work, we propose DCT-Shield, a novel approach that introduces adversarial noise in the Discrete Cosine Transform (DCT) domain of an image. This method ensures that the added noise remains highly imperceptible while offering enhanced protection compared to existing techniques, as demonstrated in Fig.~\ref{fig:teaser}. Our key contributions in this work are as follows:
\begin{itemize}
    \item We develop a novel immunization algorithm DCT-Shield that leverages the DCT domain to add highly imperceptible adversarial noise in images, while providing better protection against malicious edits compared to prior works.
    \item We demonstrate that incorporating the JPEG pipeline in the noise optimization process allows us to generate quantization-aware perturbations that remain effective even after JPEG-purification. 
    Additionally, DCT-Shield offers adaptable protection, allowing users to tune quality parameters and thresholds to balance the trade-off between purification-robustness and noise-imperceptibility.
    
    
    \item We introduce multiple DCT-Shield variants: the base version offers broad protection against edits, Mask-based DCT-Shield specializes in inpainting defense, and Y-channel DCT-Shield ensures optimal noise imperceptibility and robustness under high JPEG compression.
    \item We show that our method is parameter-efficient, requiring up to 50\% fewer parameters than pixel-space approaches. Additionally, it achieves superior results using only an encoder-based optimization, making it computationally cheaper than U-Net-based methods and provides protection against a wide range of editing models.
\end{itemize}

\section{Related Work}
\label{sec:related_work}

With advances in diffusion-based image editing models \cite{sdedit,ip2p,blended_diffusion} , it has become essential to immunize images against malicious edits. It is worth noting that \textit{all prior methods add an adversarial noise in the pixel space} of the image so as to minimize some objective, such that the immunized image when processed with an editing model, fails to produce meaningful edits. Typically, the objectives include the image's latent representation (encoder attack\footnote{In this section, `attack' refers to an adversarial attack on the editing model with an immunized image}), or the noise predicted by the U-net (diffusion attack) or their combination.
EditShield \cite{Chen_editshield} performs an encoder attack by maximizing the distance between the latents of the input and immunized images. PhotoGuard \cite{salman_photoguard} performs separate encoder and diffusion attacks by minimizing the distance between the immunized image's latent and a blank target image's latent. AdvDM \cite{liang_advdm} and MIST \cite{liang2023mist}, perform a weighted encoder and diffusion attack with a customized target image. Although they were introduced to prevent copying of art, they are also used to prevent malicious edits. DiffusionGuard \cite{choi2025diffusionguard} performs a diffusion attack to prevent malicious inpainting. Methods like \cite{Lo_2024_distraction_is_all,choi2025diffusionguard, liang_advdm, liang2023mist}, that involve diffusion attacks are typically computationally expensive due to iteratively computing gradients across large U-nets. Diff-Protect \cite{xue2024_diffprotect_sds} 
 leverages score distillation sampling to enable faster and more efficient optimization during diffusion attacks.
The authors of \cite{xue2024_diffprotect_sds} also provide convincing evidence that VAE encoders are more vulnerable to such adversarial attacks compared to U-nets in diffusion pipelines. 

However, these methods have two major shortcomings. First, the adversarial noise often appears as noticeable patterns in immunized images. Second, they lack guaranteed JPEG robustness, as immunity is typically lost when the image is compressed with JPEG. While increasing the pixel budget can improve JPEG robustness, it comes at the cost of noisy immunized images. Moreover, determining the optimal pixel budget for reliable protection is challenging.

In the next section, we discuss some preliminaries that are relevant to our proposed method.




\section{Preliminaries}
\label{sec:preliminaries}
    \begin{figure*}[htpb!]
    \centering
        \includegraphics[width=0.85\textwidth]{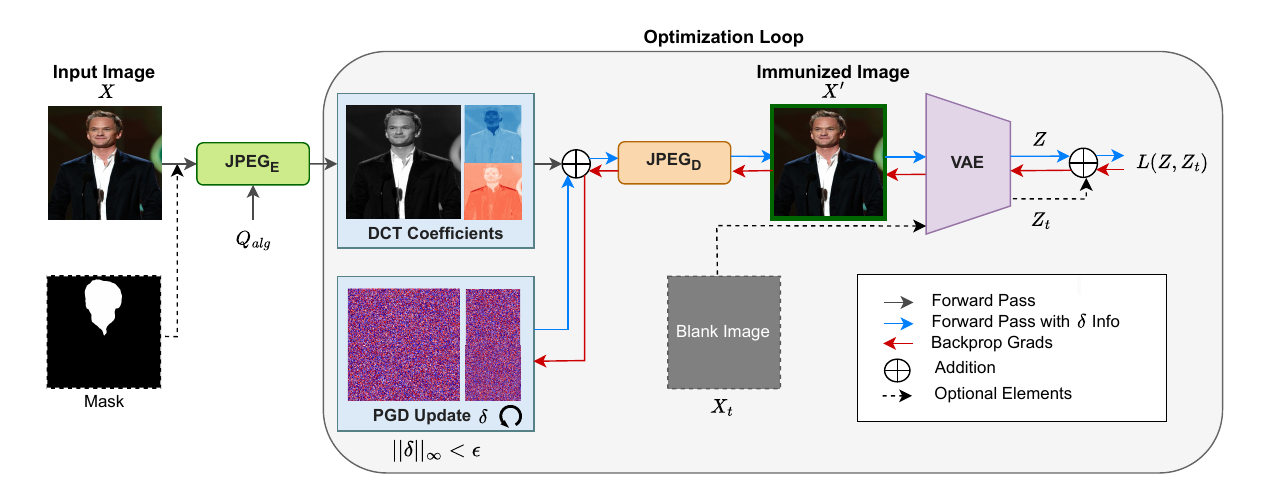}  
        \caption{ \textbf{Overview of DCT-Shield.} The algorithm takes in an input image $\mathbf{x}$, algorithm quality setting $Q_{alg}$ and perturbation bound $\epsilon$. For inpainting tasks, it also uses an additional mask as input.
        The adversarial perturbation $\delta$ is added to the input image's DCT coefficients. The adversarial (or immunized) image $\mathbf{x'}$ is passed through a VAE to obtain a corresponding latent $\mathbf{z}$, which in turn is used to define a loss function. 
        The loss is optimized using  projected gradient descent (PGD) and yields the immunized image $\mathbf{x'}$ 
        }
        \label{fig:DCTShield_method}
    \end{figure*}
\subsection{Image Editing with Latent Diffusion Models} 
\label{subsec:edit_w_ldms}
    Latent Diffusion Models (LDMs) \cite{rombach2022high} have emerged as powerful tools for image editing by leveraging a compressed latent space to perform generative tasks efficiently. Given an image $x_o\sim q(x_0)$ from the real data distribution, an LDM first uses an encoder $\mathcal{E}_{\phi}$ parameterized by $\phi$ to encode $x_0$ into a latent variable:  $z_{0}=\mathcal{E}_{\phi}(x_0)$. The forward process follows a Markovian Gaussian noise addition scheme $q(z_t|z_{t-1})=\mathcal{N}(z_t; \sqrt{1-\beta_t}z_{t-1}, \beta_tI)$, where $\beta_t$ is a variance schedule controlling the noise magnitude. Over multiple steps, the latent representation $z_t$ gradually approaches an isotropic Gaussian Distribution.
    To generate an image, the model learns to reverse the noise perturbations in the latent space. The reverse process in parameterized by a neural network $\epsilon_{\theta}(z_t, t)$ trained to estimate the noise at each step using the following objective function:  
    \begin{equation}
        \mathcal{L}(\theta) = \mathbb{E}_{z_0, t}\norm{\epsilon_{\theta}(z_t, t)-\epsilon}^2.
    \end{equation}
    Image editing tasks such as prompt-based editing and inpainting can be performed by conditioning the denoising process on additional inputs, such as text or masked regions.

\subsection{Adversarial Examples against LDMs}
\label{subsec:adv_examples_LDM}

    The key idea behind generating adversarial examples is to perturb a clean input $\mathbf{x}$ with some perturbation $\mathbf{\delta}$ such that it yields unexpected outputs from the LDM. Typically, $\mathbf{\delta}$ is added in the pixel space and an optimization problem is formulated with a specific objective. For example,
    \begin{equation}
    \label{eqn:prelim_typical_attack}
        \delta = \argmin_{\norm{\delta}_{\infty}\leq\epsilon}\mathcal{L}(f(\mathbf{x}), f(\mathbf{x + \delta})),
    \end{equation}
    where, $f(\cdot)$ can be the latent predicted by the VAE encoder 
    or the noise predicted by the U-net 
    depending on the optimization objective. There are various alternatives for the objective function $\mathcal{L}$. For example, $\mathcal{L}(u,v) = -\norm{u-v}$ is used for an untargeted adversarial optimization. The perturbation $\delta$ is constrained within an $L_{\infty}$ norm ball of radius $\epsilon$ to ensure the perturbation remains small and less perceptible. Algorithms like Projected Gradient Descent (PGD) \cite{pgd} and Fast Gradient Signed Method (FGSM) \cite{FGSM} are used to optimize $\delta$. Once obtained, it is added to the clean input image $\mathbf{x}$ to obtain an adversarial image. We note that we refer to the adversarial image as an immunized image, as its perturbations aim to protect against malicious edits.

\subsection{JPEG Algorithm}
\label{subsec:jpeg_algo}
    The JPEG algorithm \cite{jpeg} compresses images at a specified quality factor $Q_{alg}$ by removing details that are less noticeable to the human eye while maintaining perceptual similarity. 
    The JPEG encoding process consists of the following key steps:
    
    \textbf{Color Space Conversion.} JPEG encoding first converts RGB color space to YCbCr color space using a pixel-wise affine transformation.
    
    \textbf{Chroma Subsampling.} Given the reduced sensitivity of human vision to chroma details, JPEG applies chroma subsampling. This downsampling reduces the resolution of the chroma channels
    with minimal perceptual impact.
    
    \textbf{Discrete Cosine Transform (DCT).} 
    The core of JPEG compression lies in the application of the Discrete Cosine Transform (DCT). The image is divided into non-overlapping $8\times8$ pixel blocks, and each block undergoes a transformation from the spatial domain to the frequency domain. The DCT of an $8\times8$ block $p(x,y)$ is computed as:
    \begin{equation}
        \overline{\alpha}(u, v) = \frac{1}{4}C(u)C(v)\sum_{x=0}^{7}\sum_{y=0}^{7}p(x, y)g(x,u)g(y,v)
    \end{equation}
    where,
    \begin{equation}
        C(u)=\begin{cases}
            \frac{1}{\sqrt{2}}, & \text{if } u=0\\
            1, & \text{otherwise}
        \end{cases}
    \end{equation}
    
    \begin{equation}
            g(x,u) = cos\left(\frac{(2x+1)u\pi}{16}\right).\\
    \end{equation}

    \textbf{Quantization.}
    The DCT coefficients are quantized using a predefined quantization table $Q_{table}$. This step applies lossy compression by reducing the precision of high-frequency components, that are less perceptible to the human eye. The quantized coefficients $\alpha(u,v)$ are given by
    \begin{equation}
        \alpha(u, v) = \frac{\overline{\alpha}(u, v)}{Q_{table}(u, v) * f_s.}
    \end{equation}

    where, the scaling factor $f_s$ is inversely related to quality factor $Q_{alg}$.
    
    \textbf{Lossless Entropy Encoding.} 
    In the final compression step, the quantized coefficients are further compressed using Run-Length Encoding (RLE) and Huffman coding to remove redundancy.

    \textbf{JPEG Decompression}
    The JPEG decoding process reverses the encoding steps to reconstruct an approximation of the original image while maintaining perceptual quality. This involves entropy decoding, followed by de-quantization, where the quantized DCT coefficients are rescaled to approximate the original values. Then, Inverse DCT converts the frequency-domain data back into spatial-domain pixel values for each 8×8 block. This is followed by color space conversion from YCbCr back into RGB. Finally, the blocks are reassembled to reconstruct the image.

\section{Proposed Method}
\label{proposed_method}


    

    We propose DCT-Shield in an endeavor to overcome the shortcomings of prior methods (Section.~\ref{sec:related_work}) that perform optimization in the pixel space. We are motivated by the JPEG algorithm which essentially modifies image information in the DCT space to produce a JPEG image which is perceptually almost identical to the original image. Since our objective is also to introduce imperceptible modifications in the image, albeit for immunization, we develop DCT-Shield by taking advantage of the JPEG algorithm pipeline. This approach also provides a natural added advantage in terms of robustness of immunized images against JPEG-purification techniques. We note here, that although we employ the JPEG pipeline for adversarial optimization, \textit{the immunized images are free to be saved in any format.}

\subsection{Threat Model}
    We assume that a malicious editor uses a pre-trained off-the-shelf image-editing latent diffusion model (LDM) to perform an ill-intended edit on a \textit{defender's} immunized image. The \textit{defender} prepares this immunized image by adding adversarial perturbations to an original image using DCT-Shield with a pre-trained VAE in the optimization pipeline. The \textit{defender} has no prior knowledge of specific  U-net (and its corresponding weights) that the malicious editor employs nor of the prompt used to perform the mallicious edit.
    


\subsection{Problem Formulation}

    Given an input image $\mathbf{x}\in\mathbb{R}^{C\times{H}\times{W}}$ and a JPEG quality setting $Q_{alg}$, we first compute the patch-wise quantized DCT coefficients using the JPEG encode operation as follows:
    \begin{equation}
    \alpha = JPEG_{E}(\mathbf{x}; Q_{alg}), \quad \alpha(c) \in \mathbb{Z}^{n_p(c)\times8\times8},
    \end{equation}
    where, $c\in\{Y, Cb, Cr\}$ denotes the specific channel and $n_p(c)$ represents the number of patches in the channel. $JPEG_E$ denotes the sequence of compression operations described in Section~\ref{subsec:jpeg_algo} from RGB-YCbCr conversion to quantizing the DCT-coefficients.
    

    Our objective is to perturb these quantized coefficients in a way that disrupts the functionality of diffusion-based editing models. To achieve this, we target the VAE encoder within the diffusion pipeline. The key idea is to effectively optimize the perturbation $\delta$ so that the VAE encodes the adversarial image into a distorted or misleading latent representation. Therefore we first obtain an adversarial reconstruction $\mathbf{x'}$ by decompressing the perturbed quantized coefficients. Then we compute the latent representation of $\mathbf{x'}$ by passing it through a VAE encoder $\mathcal{E}$.
    Finally, the optimization objective is formulated as minimizing a loss $\mathcal{L}$ as a function of $\mathcal{E}(\mathbf{x'})$, i.e.,
    \begin{align}
        \label{eqn:DCTshield}
        \delta = \argmin_{\norm{\delta}_{\inf}\leq\epsilon} \mathcal{L}(\mathcal{E} (\mathbf{x'})), \\
        \text{ where  }\mathbf{x'} = JPEG_D(\alpha+\delta; Q_{alg}).
    \end{align}
    Here, $\delta$ denotes the collective noise across all channels, where each channel's noise $\delta(c)$ is represented as a tensor of dimensions $\mathbb{R}^{n_p(c)\times8\times8}$. $JPEG_D$ denotes the sequence of decompression operations described in Section~\ref{subsec:jpeg_algo}.
    The perturbation $\delta$ in the quantized DCT-coefficients in general is bounded by $\epsilon \in \mathbb{R}^{+}$. To ensure JPEG protection, it is crucial to induce at least one quantization level change in the quantized coefficients. Thus, the perturbation must satisfy $\epsilon\geq1$.
    We use the Projected Gradient Descent (PGD) to solve the optimization problem formulated in Eqn.~\ref{eqn:DCTshield}. We also note the fact that we add adversarial perturbations \textit{after} the quantization step. This is done to allow gradients $\nabla_\delta\mathcal{L}$ to back-propagate without having to go through the quantization function, which would otherwise zero out the gradients.  Fig.~\ref{fig:DCTShield_method} shows a schematic of the optimization process of DCT-Shield.
    
    In our experiments, we employ a norm minimization loss on the VAE latents, formulated as $\mathcal{L}(\delta) = \lVert \mathcal{E}(\mathbf{x}') \rVert_2$. This objective can also be extended to incorporate a target image $\mathbf{x}_{t}$, guiding the VAE latent representation toward a desired target latent. However, based on our empirical observations, norm minimization loss consistently delivers the best performance.
    
    Our proposed framework also allows for a diffusion-based objective, which targets the full diffusion process rather than just the VAE latents. In this approach, the loss function is defined as $\lVert f(\mathbf{x}^{'}) - \mathbf{x}_{t}\rVert_2^2$, where $f(.)$ denotes the LDM model. However, this method necessitates backpropagation through the diffusion process, making it computationally expensive. Moreover, our experiments indicate that it does not yield significant improvements over the VAE-based objective. This observation aligns with findings from Liu et al. \cite{xue2024_diffprotect_sds}, who confirm that VAEs are more susceptible to adversarial attacks than U-Nets within the diffusion process. 

    Notably, our method reduces the parameter requirement from $O(3HW)$ in pixel-space approaches to $O(3HW/2)$, matching the number of elements in the YCbCr channels after JPEG encoding. Furthermore, some specialized variants of DCT-Shield require only $O(HW)$ parameters while still maintaining strong protection against malicious edits.

\subsection{Algorithm Variants}
DCT-Shield introduces multiple variants to provide tailored protection against different types of malicious image editing while ensuring high imperceptibility. The base variant applies adversarial noise directly in the DCT domain, optimizing for robust protection across a wide range of editing tasks, including text-guided modifications, inpainting, and outpainting. However, recognizing that different types of editing and attack scenarios require specialized defenses, Mask-based DCT-Shield was specifically tailored for inpainting tasks. This variant strategically allocates noise to vulnerable regions that are more likely to be edited, enhancing the defense mechanism against localized modifications. Another specialized variant, Y-channel DCT-Shield, is designed to protect against low quality JPEG conversions by restricting adversarial perturbations to the luminance (Y) channel in the YCbCr color space. This ensures that the perturbations remain highly imperceptible 
while maintaining robust protection. 
Therefore, DCT-Shield's variants provide a flexible defense, enabling users to balance noise-imperceptibility, protection-robustness, and efficiency based on their needs. Please refer to the detailed algorithm in Section A of the supplementary material.


\section{Experiments}

\subsection{Benchmark Dataset}
\label{subsec:datasets}
    We use 150 samples from the OmniEdit \cite{omniedit} dataset covering a variety of edit tasks like object addition, removal, replacement and  modifications of attributes and environment. For the inpainting task, we constructed a dataset consisting of 56 samples. These samples include images sourced from the PPR10K dataset \cite{ppr10k} as well as portrait images collected from the web. Further details and dataset samples are shown in Section B of the supplementary material.
    
\subsection{Baselines}
\label{subsec:baselines}
    We compare our method, DCT-Shield, with various baseline methods, namely PhotoGuard \cite{salman_photoguard}, MIST \cite{liang2023mist}, AdvDM \cite{liang_advdm}, and Diff-Protect (SDS) \cite{xue2024_diffprotect_sds}. For the image inpainting task, we additionally compare DCT-Shield with the state-of-the-art (SOTA) method DiffusionGuard \cite{choi2025diffusionguard}.
    
\subsection{Edit Models and Evaluation Metrics}
\label{subsec:eval_metrics}
   For our experiments, we utilize InstructPix2Pix (IP2P) \cite{ip2p}, a widely used diffusion-based editing model, for instruction-based image editing. For the inpainting task, we employ the Stable Diffusion Inpainting 1.0 model \cite{sd}. Additionally, we demonstrate the cross-model transferability of our approach in Section D.5 of the supplementary material.

    We evaluate three key aspects while comparing our method with the baseline methods:
    \begin{enumerate}[label=\roman*.]
        \item Perceivability of the added adversarial noise, 
        \item Semantic difference between the edits of the original and immunized images for evaluating the protection quality,
        \item Semantic difference between the edits of the original and 
        purified immunized images for evaluating the robustness to perturbation purification techniques.
    \end{enumerate}
    We quantify semantic differences using LPIPS \cite{lpips}, FID \cite{fid}, PSNR, SSIM \cite{ssim}, and VIFp \cite{vifp}. For inpainting, we also report CLIPScore \cite{clipscore}. Additionally, we conduct extensive human evaluations for both editing and inpainting. Further details are in the Section C of the supplementary.

\subsection{Implementation Details}
\label{subsec:implementation}
    We run DCT-Shield with the following default settings. Immunization quality is set to $Q_{alg} = 0.95$ 
    and the adversarial perturbation in the quantized DCT-coefficients is bound by $\epsilon = 1$ across all channels. The step size $\gamma = 0.1$ The algorithm is run for 1000 iterations. We work with images of resolution $512\times512$. We clearly state different settings in the text if they are changed for an ablation or experiment.
    
    \begin{figure*}[t]
    \centering
        \includegraphics[width=0.8\textwidth]{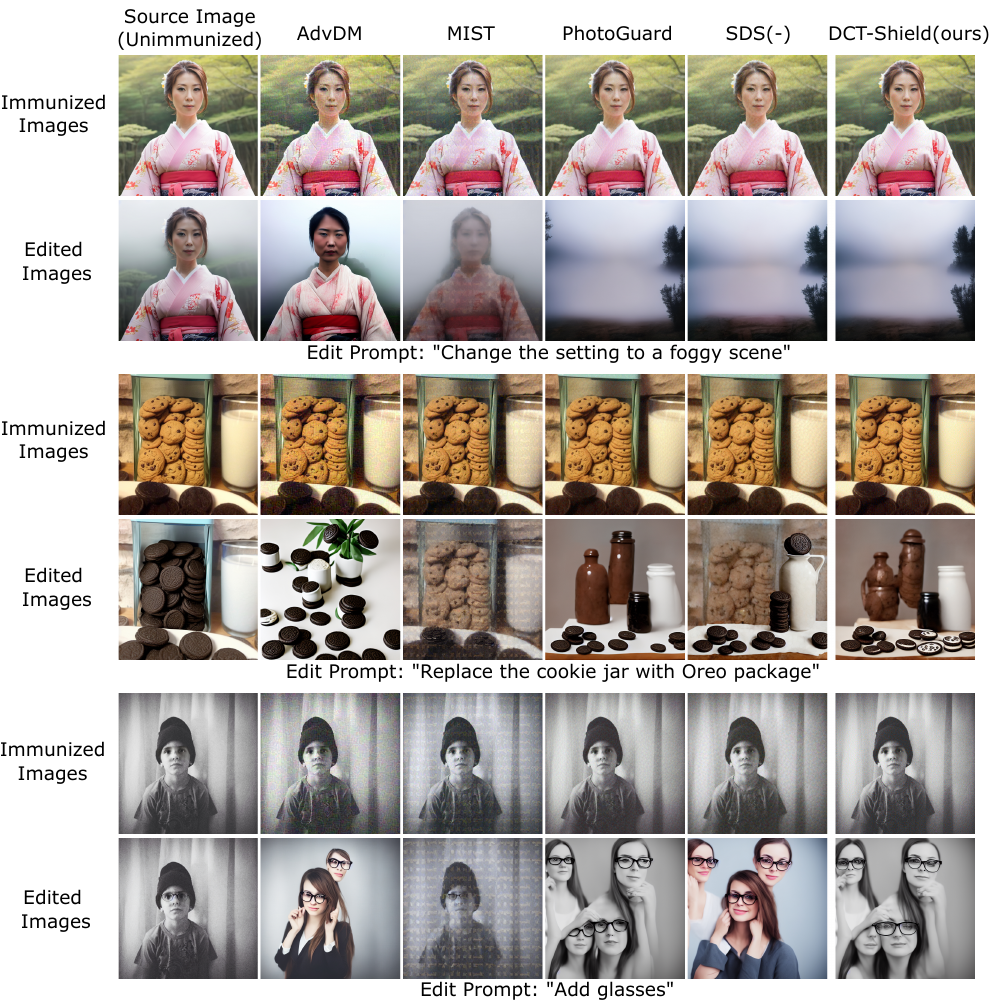}  
        \caption{\textbf{Qualitative analysis comparing DCT-Shield with baseline methods.} For each example, the first row presents the original input (source) image alongside the corresponding immunized images. The second row illustrates the edited version of the original image, followed by the edited immunized images across different baselines. DCT-Shield exhibits comparable or superior resistance to edits while maintaining significantly lower perceptible noise in the immunized images across various editing tasks. }
        \label{fig:qualitative_comparison}
    \end{figure*}
\section{Results}

\subsection{Immunization Quality}

   We assess DCT-Shield's immunization quality by evaluating \textit{(i)} the perceptibility of added adversarial noise and \textit{(ii)} its effectiveness in preventing malicious edits. For comparison, baseline methods use a standard pixel budget ($L_{\infty}$-norm) of $16/255$ pixels, while DCT-Shield operates at default settings ($Q_{alg}=0.95$, $\epsilon=1$) as described in Sec.~\ref{subsec:implementation}. Unlike baselines, DCT-Shield does not impose an explicit pixel budget, but this is not a limitation, as RGB space norms do not reliably reflect human visual perception \cite{sharif2018suitability, zhao_color_loss}. Thus, pixel budgets are not a suitable metric for noise perceptibility.
   
    \textbf{Qualitative results.} Figure~\ref{fig:qualitative_comparison} showcases qualitative comparisons across various editing tasks. DCT-Shield introduces significantly less perceptible adversarial noise while providing strong protection against edits, including environment changes (first row), object replacement (second row), and object addition (third row). Notably, edits on DCT-Shield’s immunized images fail to preserve key semantics, ensuring effective protection, whereas baselines sometimes fall short. More qualitative results across edit-types are shown in Section D of the supplementary material.

    \textbf{Quantitative results.} We also show quantitative results in Table~\ref{tab:edit_protection} to concretely showcase DCT-Shield's immunization quality. DCT-Shield typically outperforms baseline methods across various metrics, in terms of both edit-protection and noise imperceptibility. Notably, in cases where DCT-Shield does not provide the best perception levels, it indeed does so in the corresponding protection metric and vice versa. This indicates a trade-off between the two aspects and is discussed in the next paragraph. Additionally, human evaluation scores indicate a clear preference for DCT-Shield’s immunized images and its ability to safeguard against edits. Details of human evaluation studies are presented in 
    Section C.2 of the supplementary material. 

\begin{figure}[h]
    \centering
    \includegraphics[width=0.3\textwidth]{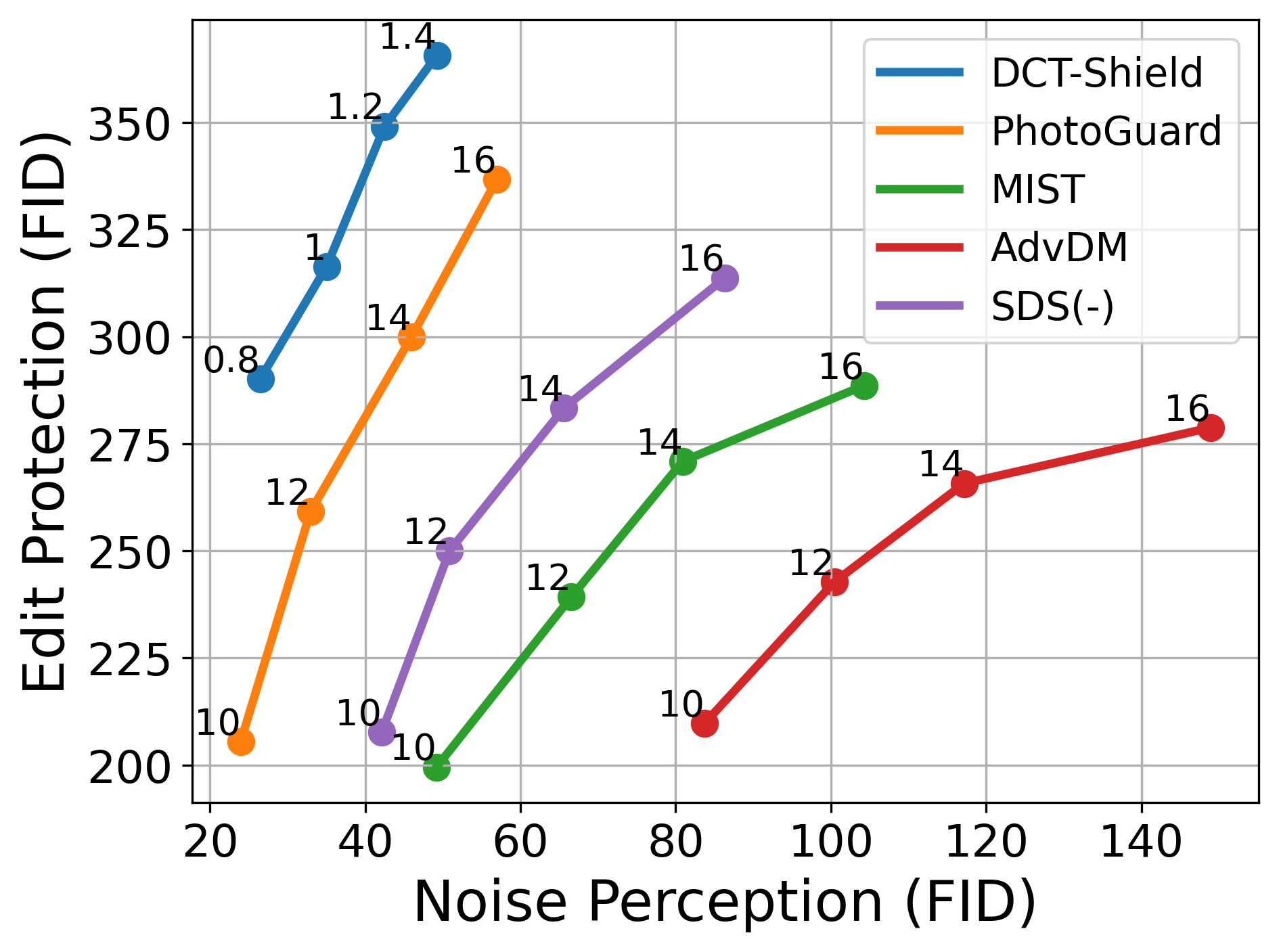}
    \caption{\textbf{Comparison of the edit-protection vs. noise perception trade-off with baselines.} Edit-protection is measured by the FID score (higher is better) between edited versions of input and immunized images. Noise perception is assessed by the FID score (lower is better) between input and immunized images. The numbers on the dots are perturbation limits $\epsilon$. DCT-Shield achieves a superior trade-off, offering comparable edit-protection while maintaining significantly lower noise perceptibility.} 
    \label{fig:protection_vs_noise_perception}
    
\end{figure}

    \textbf{Protection vs. noise-imperceptibility trade-off.} Comparing perturbations in DCT coefficients to those in pixel space is challenging. To ensure fairness, we evaluate DCT-Shield at different coefficient perturbation limits $\epsilon = [0.8, 1, 1.2, 1.4]$ and baselines at pixel-space budgets of $10,12,14$ and $16$ ($/255$). Fig.~\ref{fig:protection_vs_noise_perception} illustrates the trade-off between edit protection and noise perceptibility across immunization methods. Noise perceptibility is measured by the FID score between input and immunized images (lower is better), while edit protection is evaluated by the FID score between edited versions (higher is better). DCT-Shield achieves a superior balance, providing strong edit protection with minimal perceptible noise.
    
\begin{table*}[!ht]
\centering
 \resizebox{0.98\textwidth}{!}{ 
 \renewcommand{\arraystretch}{1.2}
\begin{tabular}{lcccccccccccc}
\hline
                                      & \multicolumn{6}{c}{\textbf{Noise Perception}}                                                                                    & \multicolumn{6}{c}{\textbf{Edit Protection}}                                                                \\ \hline
\multicolumn{1}{l|}{\textbf{Methods}} & \textbf{LPIPS}$\downarrow$  & \textbf{FID}$\downarrow$    & \textbf{SSIM}$\uparrow$  & \textbf{PSNR}$\uparrow$   & \textbf{VIFp}$\uparrow$  & \multicolumn{1}{c|}{\textbf{HEval}$\uparrow$} & \textbf{LPIPS}$\uparrow$ & \textbf{FID}$\uparrow$     & \textbf{SSIM}$\downarrow$  & \textbf{PSNR}$\downarrow$   & \textbf{VIFp}$\downarrow$  & \textbf{HEval}$\uparrow$ \\ \hline
\multicolumn{1}{l|}{AdvDM \cite{liang_advdm}}            & 0.353          & 148.890         & 0.750          & 27.114          & 0.316          & \multicolumn{1}{c|}{2.35}           & 0.561          & 278.754          & 0.498          & 13.194          & 0.075          & 3.44           \\
\multicolumn{1}{l|}{Mist \cite{liang2023mist}}             & 0.362          & 104.265         & 0.730          & 26.620          & 0.344          & \multicolumn{1}{c|}{2.12}           & 0.534          & 288.61           & 0.504          & 16.550          & 0.087          & 2.45           \\
\multicolumn{1}{l|}{PhotoGuard \cite{salman_photoguard}}  & 0.284          & 57.782          & \textbf{0.829} & \textbf{28.323} & 0.543          & \multicolumn{1}{c|}{3.96}           & 0.679          & \textbf{336.736} & 0.450          & 12.546          & 0.059          & 4.16           \\
\multicolumn{1}{l|}{SDS(-) \cite{xue2024_diffprotect_sds}}            & 0.335          & 86.359          & 0.728          & 27.838          & 0.437          & \multicolumn{1}{c|}{3.65}           & 0.681          & 313.735          & 0.452          & 12.699          & 0.061          & 4.02           \\
\multicolumn{1}{l|}{DCT-Shield}        & \textbf{0.267} & \textbf{35.023} & 0.822          & 27.612          & \textbf{0.776} & \multicolumn{1}{c|}{\textbf{4.22}}  & \textbf{0.684} & 316.363          & \textbf{0.448} & \textbf{12.247} & \textbf{0.058} & \textbf{4.35}  \\ \hline
\end{tabular}
}
\caption{\textbf{Quantitative comparison of DCT-Shield with baseline methods}. Noise Perception indicates the similarity between the immunized image and the original image, while Edit Protection quantifies the difference between the immunized edit and the clean edit. We denote the human evaluation results as HEval.}
\label{tab:edit_protection}
\end{table*}

\begin{table}[ht]
\resizebox{0.95\linewidth}{!}{
\renewcommand{\arraystretch}{1.2}
\begin{tabular}{l|ccccccc}
\hline
\textbf{Method} & \textbf{LPIPS}$\uparrow$ & \textbf{FID}$\uparrow$     & \textbf{SSIM}$\downarrow$ & \textbf{PSNR}$\downarrow$  & \textbf{VIFp}$\downarrow$  & \textbf{CLIP}$\downarrow$  & \textbf{Human Eval} \\ \hline
Unprotected     & NA             & NA               & NA            & NA             & NA             & 0.704          & NA                  \\
AdvDM \cite{liang_advdm}          & 0.421          & 170.389          & 0.634         & 16.770         & 0.319          & 0.706          & 2.32                \\
Mist \cite{liang2023mist}           & 0.482          & 187.113          & 0.564         & 15.688         & 0.290          & 0.693          & 2.84                \\
PhotoGuard \cite{salman_photoguard}     & 0.506          & 180.315          & 0.576         & 16.669         & 0.272          & 0.682          & 3.36                \\
SDS(-) \cite{xue2024_diffprotect_sds}         & 0.473          & 178.922          & 0.580         & 16.60          & 0.297          & 0.70           & 2.58                \\
DiffusionGuard \cite{choi2025diffusionguard}  & 0.518          & 194.932          & 0.584         & 16.459         & 0.272          & \textbf{0.664} & 3.96                \\
DCT-Shield      & \textbf{0.547} & \textbf{199.082} & \textbf{0.531}         & \textbf{15.95} & \textbf{0.261} & 0.674          & \textbf{4.12}                \\ \hline
\end{tabular}
}
\caption{\textbf{Quantitative results on the inpainting task.} Comparison against baseline methods for the inpainting task. DCT-Shield demonstrates superior protection compared to baselines.}
\label{tab:inp_protection}
\end{table}

    \begin{figure*}[t]
    \centering
        \includegraphics[width=0.75\textwidth,keepaspectratio]{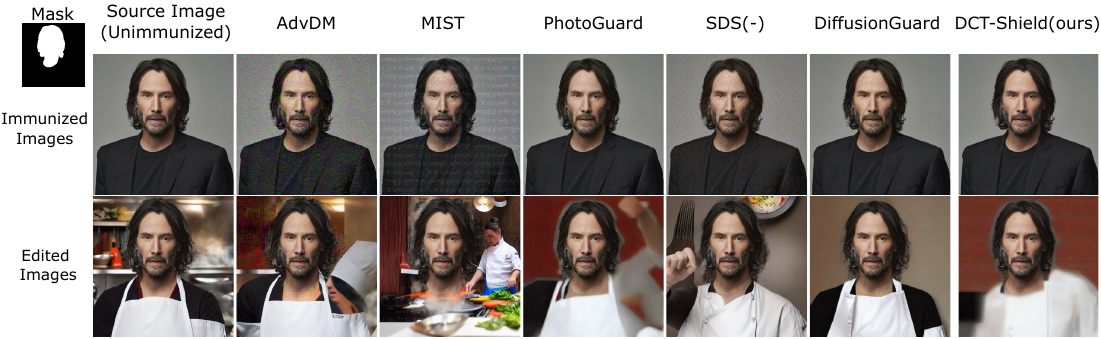}  
        \caption{\textbf{Qualitative results for inpainting tasks.} The first row presents the input (source) image alongside immunized images generated by different methods. The second row displays the inpainting edit of the input image, followed by the corresponding edits of the immunized images. The edit prompt used is: "chef cooking in a restaurant". Compared to baseline methods, DCT-Shield provides stronger edit-protection with less perceptible noise in the immunized image.}
        \label{fig:inpainting_qual}
    \end{figure*}
    
\begin{figure}[h]
    \centering
            \includegraphics[width=0.42\textwidth,keepaspectratio]{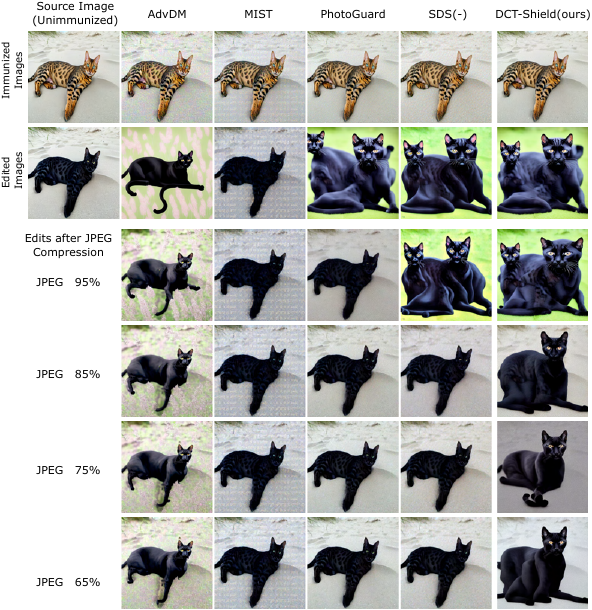}  
    \caption{\textbf{Qualitative results on JPEG robustness.} The first two rows show the original and immunized images with their corresponding edits. Subsequent rows show edits on purified images, with each row corresponding to a given JPEG quality. The edit prompt is "Make the Bengal cat black". DCT-Shield retains edit-protection abilities across a wide range of quality levels, while most baseline methods fail, especially at low JPEG quality values.}
    \label{fig:jpeg_robustness}
\end{figure}

\subsection{Immunization Against Malicious Inpainting}
    Defense against inpainting attacks poses a significant challenge, as an attacker can mask out a sensitive region from an image and seamlessly blend it into a new background. Effective protection requires adding targeted perturbations to these sensitive regions to disrupt inpainting models. To mitigate this threat, we employ DCT-Shield with a quality parameter of $Q_{alg}=0.9$, applying masked noise to the YCbCr channels. This configuration ensures strong inpainting protection while preserving the imperceptibility of the added noise. All baseline methods are run with a pixel budget of $16$. We evaluate our approach against baseline methods in Table \ref{tab:inp_protection}, where results demonstrate that DCT-Shield outperforms or is at least comparable across all evaluation metrics. A qualitative comparison of inpainting results is presented in Figure \ref{fig:inpainting_qual}, illustrating that our method provides stronger protection by effectively preventing the attacker's intent. Notably, we achieve these results using only encoder-based optimization, whereas several baselines rely on diffusion-based optimization, which is computationally more intensive. Additional qualitative results for this task are shown in Section D of the supplementary material.



        \begin{figure}[t]
            \centering
            \includegraphics[width=0.28\textwidth, keepaspectratio]{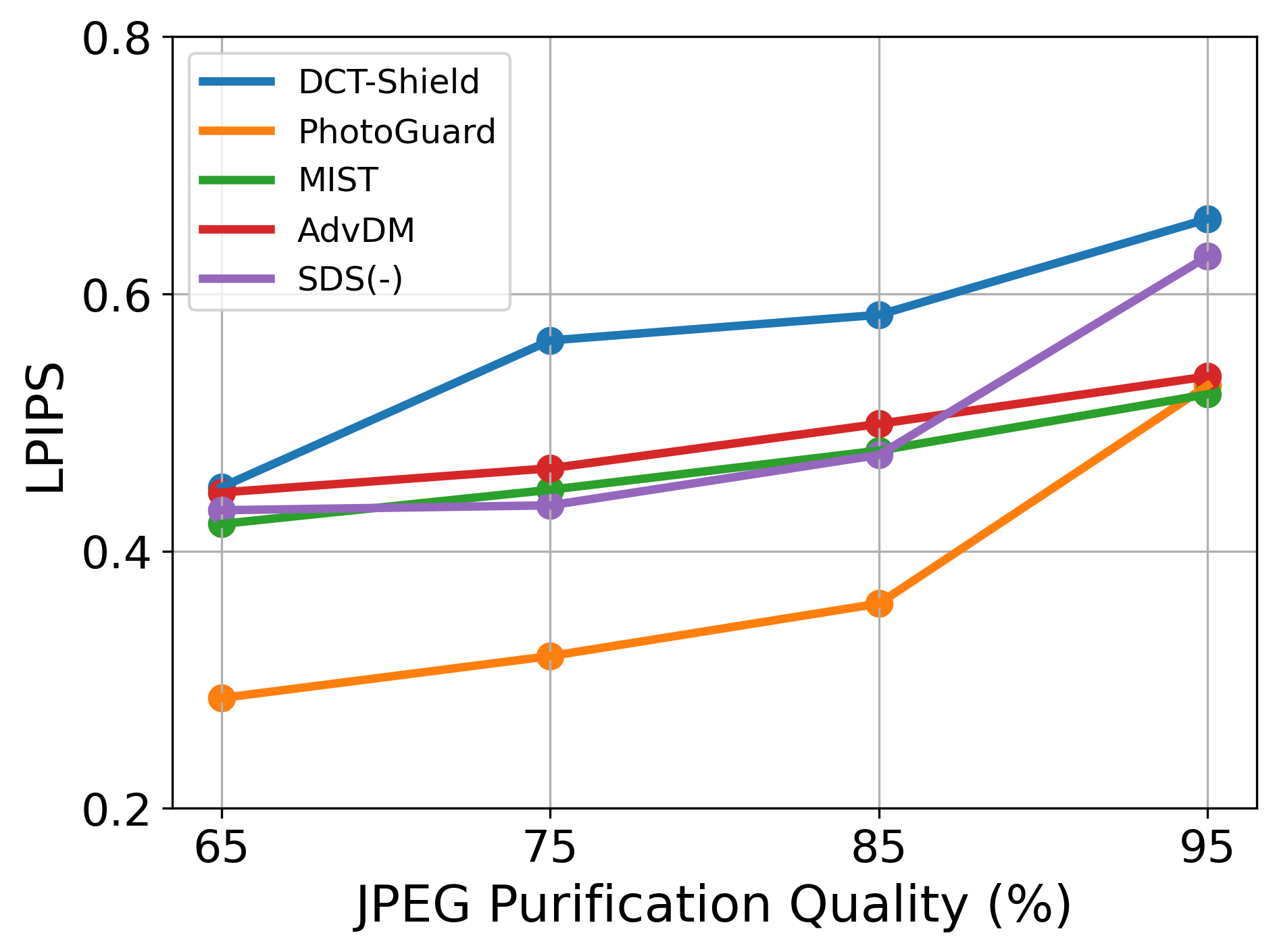}
            \caption{\textbf{Comparison of robustness against JPEG-purification}. LPIPS($\uparrow$) score is computed between edits of original and purified images. DCT-Shield maintains higher score across varying levels of JPEG purification, indicating stronger robustness.}
            \label{fig:jpeg_robust_annalysis}
        \end{figure}

\subsection{Robustness to Purification Techniques}
Adversarial noise purification techniques aim to remove immunization perturbations while preserving data integrity. Common methods include JPEG compression and crop \& resize, while advanced approaches like AdvClean\footnote{\url{https://github.com/lllyasviel/AdverseCleaner}} use image filtering. Malicious editors may exploit these techniques to reverse immunization. This section evaluates DCT-Shield’s robustness against such purification methods, comparing it to baseline approaches. Results for purification methods like Gaussian noising, Impress\cite{Impress} and noisy upscaling are shown in Section D.3 of the supplementary.

    \textbf{JPEG Conversion.} We evaluate edit protection after JPEG purification at quality levels 65, 75, 85, and 95 percent. Fig.~\ref{fig:jpeg_robustness} shows qualitative results with DCT-Shield (only-Y) at $Q_{alg}=0.85$. The figure highlights DCT-Shield’s superior capability of retaining image immunity despite JPEG conversions. This is evident from the semantics of the cat in the last column (DCT-Shield) vs. other columns (baselines) across all JPEG quality values.
    
    Quantitative results in Fig.~\ref{fig:jpeg_robust_annalysis} confirm DCT-Shield’s resilience using LPIPS, with similar trends observed in other metrics. Unlike pixel-space methods, DCT-Shield integrates JPEG pipeline in noise optimization, ensuring that images immunized generated with $Q_{alg} = q$ remain largely unaffected by JPEG purification at quality levels exceeding $q$. Furthermore, this enables defenders to fine-tune $Q_{alg}$ to optimize the trade-off between robustness range and noise perceptibility. Notably, such adaptability is not attainable through pixel-space optimization methods. Qualitative results showing range robustness are shown in Section D.4 of the supplementary material.

\begin{figure}[h]
            \centering
            \includegraphics[width=0.42\textwidth, keepaspectratio]{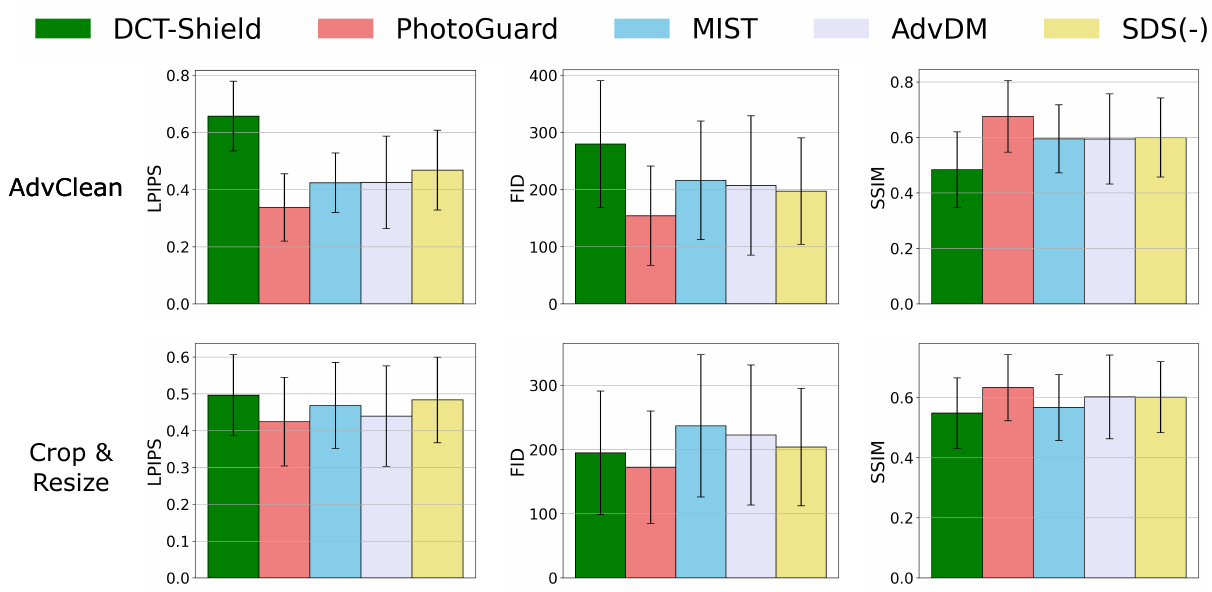}
            \caption{\textbf{Robustness against AdvClean and crop-and-resize}. Comparison of DCT-Shield with baseline methods in terms of
            LPIPS($\uparrow$) , FID($\uparrow$) and SSIM($\downarrow$) between edits of original  and purified images. DCT-Shield demonstrates overall better robustness against both purification methods.}
            \label{fig:transformation_robust}
        \end{figure}

    \textbf{Crop and Resize.}
Following the approach in \cite{liang_advdm, choi2025diffusionguard}, immunized images were cropped by 64 pixels at the boundaries and then resized to their original dimensions $512 \times 512$. The edits of these cropped and resized immunized images were compared to the edits of the original input. As shown in Fig.~\ref{fig:transformation_robust}, DCT-Shield demonstrates greater robustness to this crop-and-resize purification method.

    \textbf{AdvClean.}
    We further analyze the edits of immunized images after purification using AdvClean, comparing them to edits of the original images. As shown in Fig.~\ref{fig:transformation_robust}, DCT-Shield exhibits greater robustness to AdvClean purification in terms of LPIPS and SSIM while maintaining performance comparable to baseline methods in terms of FID.


\section{Conclusion}

In this work, we present DCT-Shield - a novel image immunization technique that operates in the Discrete Cosine Transform (DCT) domain to embed imperceptible yet effective perturbations, protecting images from malicious edits. By leveraging the JPEG pipeline during optimization, DCT-Shield achieves robustness against a wide range of purification methods while preserving high visual fidelity. It also allows users to control the balance between protection-robustness and noise-imperceptibility through adjustable parameters. We further introduced multiple DCT-Shield variants, designed to address task-specific requirements and improve parameter efficiency. Extensive qualitative and quantitative evaluations demonstrate that DCT-Shield outperforms prior methods in both edit protection and imperceptibility. DCT-Shield employs significantly fewer parameters than pixel-space approaches. Furthermore, by utilizing only a VAE in the optimization process, it remains computationally efficient and independent of specific U-Net architectures across different editing models. This makes DCT-Shield a practical and scalable solution for real-world image security. Future research can explore extending DCT-Shield to video data and enhancing its robustness against emerging purification techniques.
{
    \small
    \bibliographystyle{ieeenat_fullname}
    \bibliography{main}

\begin{thebibliography}{25}
\providecommand{\natexlab}[1]{#1}
\providecommand{\url}[1]{\texttt{#1}}
\expandafter\ifx\csname urlstyle\endcsname\relax
  \providecommand{\doi}[1]{doi: #1}\else
  \providecommand{\doi}{doi: \begingroup \urlstyle{rm}\Url}\fi

\bibitem[Avrahami et~al.(2022)Avrahami, Lischinski, and Fried]{blended_diffusion}
Omri Avrahami, Dani Lischinski, and Ohad Fried.
\newblock Blended diffusion for text-driven editing of natural images.
\newblock In \emph{2022 IEEE/CVF Conference on Computer Vision and Pattern Recognition (CVPR)}, page 18187–18197. IEEE, 2022.

\bibitem[Brooks et~al.(2022)Brooks, Holynski, and Efros]{ip2p}
Tim Brooks, Aleksander Holynski, and Alexei~A Efros.
\newblock Instructpix2pix: Learning to follow image editing instructions.
\newblock \emph{arXiv preprint arXiv:2211.09800}, 2022.

\bibitem{Impress}
Bochuan Cao, Changjiang Li, Ting Wang, Jinyuan Jia, Bo Li, and Jinghui Chen.
\newblock \textit{IMPRESS: Evaluating the Resilience of Imperceptible Perturbations Against Unauthorized Data Usage in Diffusion-Based Generative AI}.
\newblock In \emph{Advances in Neural Information Processing Systems}, pages 10657--10677. Curran Associates, Inc., 2023.
\newblock URL: \url{https://proceedings.neurips.cc/paper_files/paper/2023/file/222dda29587fbc2979ca99fd5ed00735-Paper-Conference.pdf}.

\bibitem[Chen et~al.(2025)Chen, Jin, Liu, Chen, Wang, and Sun]{Chen_editshield}
Ruoxi Chen, Haibo Jin, Yixin Liu, Jinyin Chen, Haohan Wang, and Lichao Sun.
\newblock Editshield: Protecting unauthorized image editing by instruction-guided diffusion models.
\newblock In \emph{Computer Vision -- ECCV 2024}, pages 126--142, Cham, 2025. Springer Nature Switzerland.

\bibitem[Choi et~al.(2025)Choi, Lee, Jeong, Xie, Shin, and Lee]{choi2025diffusionguard}
June~Suk Choi, Kyungmin Lee, Jongheon Jeong, Saining Xie, Jinwoo Shin, and Kimin Lee.
\newblock Diffusionguard: A robust defense against malicious diffusion-based image editing.
\newblock In \emph{The Thirteenth International Conference on Learning Representations}, 2025.

\bibitem[Goodfellow et~al.(2015)Goodfellow, Shlens, and Szegedy]{FGSM}
Ian~J. Goodfellow, Jonathon Shlens, and Christian Szegedy.
\newblock Explaining and harnessing adversarial examples, 2015.

\bibitem[Hessel et~al.(2022)Hessel, Holtzman, Forbes, Bras, and Choi]{clipscore}
Jack Hessel, Ari Holtzman, Maxwell Forbes, Ronan~Le Bras, and Yejin Choi.
\newblock Clipscore: A reference-free evaluation metric for image captioning, 2022.

\bibitem[Heusel et~al.(2018)Heusel, Ramsauer, Unterthiner, Nessler, and Hochreiter]{fid}
Martin Heusel, Hubert Ramsauer, Thomas Unterthiner, Bernhard Nessler, and Sepp Hochreiter.
\newblock Gans trained by a two time-scale update rule converge to a local nash equilibrium, 2018.

\bibitem[Liang and Wu(2023)]{liang2023mist}
Chumeng Liang and Xiaoyu Wu.
\newblock Mist: Towards improved adversarial examples for diffusion models.
\newblock \emph{arXiv preprint arXiv:2305.12683}, 2023.

\bibitem[Liang et~al.(2023)Liang, Wu, Hua, Zhang, Xue, Song, Xue, Ma, and Guan]{liang_advdm}
Chumeng Liang, Xiaoyu Wu, Yang Hua, Jiaru Zhang, Yiming Xue, Tao Song, Zhengui Xue, Ruhui Ma, and Haibing Guan.
\newblock Adversarial example does good: Preventing painting imitation from diffusion models via adversarial examples.
\newblock In \emph{Proceedings of the 40th International Conference on Machine Learning}, pages 20763--20786. PMLR, 2023.

\bibitem[Liang et~al.(2021)Liang, Zeng, Cui, Xie, and Zhang]{ppr10k}
Jie Liang, Hui Zeng, Miaomiao Cui, Xuansong Xie, and Lei Zhang.
\newblock Ppr10k: A large-scale portrait photo retouching dataset with human-region mask and group-level consistency.
\newblock In \emph{Proceedings of the IEEE Conference on Computer Vision and Pattern Recognition}, 2021.

\bibitem[Liu et~al.(2023)Liu, Lau, and Chellappa]{liu2023diffprotect}
Jiang Liu, Chun~Pong Lau, and Rama Chellappa.
\newblock Diffprotect: Generate adversarial examples with diffusion models for facial privacy protection.
\newblock \emph{arXiv preprint arXiv:2305.13625}, 2023.

\bibitem[Lo et~al.(2024)Lo, Yeo, Shuai, and Cheng]{Lo_2024_distraction_is_all}
Ling Lo, Cheng~Yu Yeo, Hong-Han Shuai, and Wen-Huang Cheng.
\newblock Distraction is all you need: Memory-efficient image immunization against diffusion-based image editing.
\newblock In \emph{Proceedings of the IEEE/CVF Conference on Computer Vision and Pattern Recognition (CVPR)}, pages 24462--24471, 2024.

\bibitem[Madry et~al.(2019)Madry, Makelov, Schmidt, Tsipras, and Vladu]{pgd}
Aleksander Madry, Aleksandar Makelov, Ludwig Schmidt, Dimitris Tsipras, and Adrian Vladu.
\newblock Towards deep learning models resistant to adversarial attacks, 2019.

\bibitem[Meng et~al.(2022)Meng, He, Song, Song, Wu, Zhu, and Ermon]{sdedit}
Chenlin Meng, Yutong He, Yang Song, Jiaming Song, Jiajun Wu, Jun-Yan Zhu, and Stefano Ermon.
\newblock Sdedit: Guided image synthesis and editing with stochastic differential equations, 2022.

\bibitem[Rombach et~al.(2021)Rombach, Blattmann, Lorenz, Esser, and Ommer]{sd}
Robin Rombach, Andreas Blattmann, Dominik Lorenz, Patrick Esser, and Björn Ommer.
\newblock High-resolution image synthesis with latent diffusion models, 2021.

\bibitem[Rombach et~al.(2022)Rombach, Blattmann, Lorenz, Esser, and Ommer]{rombach2022high}
Robin Rombach, Andreas Blattmann, Dominik Lorenz, Patrick Esser, and Bj{\"o}rn Ommer.
\newblock High-resolution image synthesis with latent diffusion models.
\newblock In \emph{Proceedings of the IEEE/CVF conference on computer vision and pattern recognition}, pages 10684--10695, 2022.

\bibitem[Salman et~al.(2023)Salman, Khaddaj, Leclerc, Ilyas, and M\k{a}dry]{salman_photoguard}
Hadi Salman, Alaa Khaddaj, Guillaume Leclerc, Andrew Ilyas, and Aleksander M\k{a}dry.
\newblock Raising the cost of malicious ai-powered image editing.
\newblock In \emph{Proceedings of the 40th International Conference on Machine Learning}. JMLR.org, 2023.

\bibitem[Sharif et~al.(2018)Sharif, Bauer, and Reiter]{sharif2018suitability}
Mahmood Sharif, Lujo Bauer, and Michael~K Reiter.
\newblock On the suitability of lp-norms for creating and preventing adversarial examples.
\newblock In \emph{Proceedings of the IEEE conference on computer vision and pattern recognition workshops}, pages 1605--1613, 2018.

\bibitem[Sheikh and Bovik(2006)]{vifp}
Hamad~R. Sheikh and Alan~C. Bovik.
\newblock Image information and visual quality.
\newblock \emph{IEEE Transactions on Image Processing}, 15\penalty0 (2):\penalty0 430--444, 2006.

\bibitem[Wallace(1992)]{jpeg}
G.K. Wallace.
\newblock The jpeg still picture compression standard.
\newblock \emph{IEEE Transactions on Consumer Electronics}, 38\penalty0 (1):\penalty0 xviii--xxxiv, 1992.

\bibitem[Wang et~al.(2004)Wang, Bovik, Sheikh, and Simoncelli]{ssim}
Zhou Wang, A.C. Bovik, H.R. Sheikh, and E.P. Simoncelli.
\newblock Image quality assessment: from error visibility to structural similarity.
\newblock \emph{IEEE Transactions on Image Processing}, 13\penalty0 (4):\penalty0 600--612, 2004.

\bibitem[Wei et~al.(2024)Wei, Xiong, Ren, Du, Zhang, and Chen]{omniedit}
Cong Wei, Zheyang Xiong, Weiming Ren, Xinrun Du, Ge Zhang, and Wenhu Chen.
\newblock Omniedit: Building image editing generalist models through specialist supervision.
\newblock \emph{arXiv preprint arXiv:2411.07199}, 2024.

\bibitem[Xue et~al.(2024)Xue, Liang, Wu, and Chen]{xue2024_diffprotect_sds}
Haotian Xue, Chumeng Liang, Xiaoyu Wu, and Yongxin Chen.
\newblock Toward effective protection against diffusion-based mimicry through score distillation.
\newblock In \emph{The Twelfth International Conference on Learning Representations}, 2024.

\bibitem[Zhang et~al.(2018)Zhang, Isola, Efros, Shechtman, and Wang]{lpips}
Richard Zhang, Phillip Isola, Alexei~A Efros, Eli Shechtman, and Oliver Wang.
\newblock The unreasonable effectiveness of deep features as a perceptual metric.
\newblock In \emph{CVPR}, 2018.

\bibitem[Zhao et~al.(2020)Zhao, Liu, and Larson]{zhao_color_loss}
Zhengyu Zhao, Zhuoran Liu, and Martha Larson.
\newblock Towards large yet imperceptible adversarial image perturbations with perceptual color distance.
\newblock In \emph{2020 IEEE/CVF Conference on Computer Vision and Pattern Recognition (CVPR)}, pages 1036--1045, 2020.

\end{thebibliography}
}

\end{document}